\documentclass[letterpaper]{article} 
\usepackage{aaai24}  
\usepackage{times}  
\usepackage{helvet}  
\usepackage{courier}  
\usepackage[hyphens]{url}  
\usepackage{graphicx} 
\usepackage{natbib}  
\usepackage{caption} 
\usepackage{algorithm}
\usepackage{algorithmic}
\usepackage{bm}
\usepackage{amsfonts}

\usepackage{newfloat}
\usepackage{listings}
\usepackage{amsmath}
\usepackage{amsthm}

\DeclareCaptionStyle{ruled}{labelfont=normalfont,labelsep=colon,strut=off} 
\floatstyle{ruled}
\newfloat{listing}{tb}{lst}{}
\floatname{listing}{Listing}
\title{A Variational Autoencoder for Neural Temporal Point Processes with Dynamic Latent Graphs}
\author{
   Sikun Yang\textsuperscript{\rm 1, 2, 3},
    Hongyuan Zha\textsuperscript{\rm 4, 5}
}
\affiliations{
\textsuperscript{\rm 1} School of Computing and Information Technology, Great Bay University, 523000 Dongguan, China\\
\textsuperscript{\rm 2} Great Bay Institute for Advanced Study, Great Bay University\\
\textsuperscript{\rm 3} Dongguan Key Laboratory for Data Science and Intelligent Medicine, Great Bay University\\
\textsuperscript{\rm 4} Shenzhen Institute of Artificial Intelligence and Robotics for Society\\
\textsuperscript{\rm 5} The Chinese University of Hong Kong, Shenzhen\\
{sikunyang@gbu.edu.cn, zhahy@cuhk.edu.cn}%
}

\usepackage{bibentry}

\begin{document}

\maketitle

\begin{abstract}
Continuously-observed event occurrences, often exhibit self- and mutually-exciting effects, which can be well modeled using temporal point processes.  
Beyond that, these event dynamics may also change over time, with certain periodic trends. 
We propose a novel variational auto-encoder to capture such a mixture of temporal dynamics. 
More specifically, the whole time interval of the input sequence is partitioned into a set of sub-intervals. The event dynamics are assumed to be {stationary} within each sub-interval, but could be changing across those sub-intervals. In particular, we use a sequential latent variable model to learn a dependency graph between the observed dimensions, for each sub-interval. The model predicts the future event times, by using the learned dependency graph to remove the non-contributing influences of past events. By doing so, the proposed model demonstrates its higher accuracy in predicting inter-event times and event types for several real-world event sequences, compared with existing state of the art neural point processes. 
\end{abstract}

\section{Introduction}
\begin{figure}[t!]
  \centering
      \includegraphics[width=8.2cm,height=10cm, keepaspectratio]{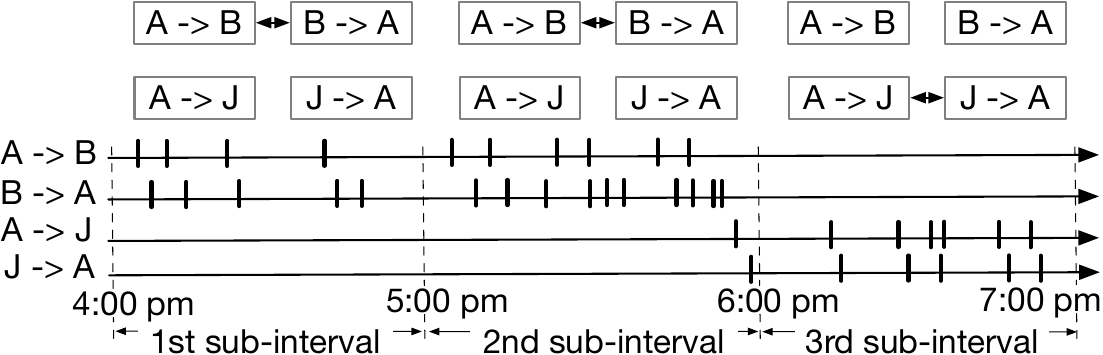}
\caption{An example illustrates the email interactions between three individuals, \emph{Alice} (A), \emph{Bob} (B), \emph{Jane} (J). During working time, \emph{Alice} sends emails (black sticks) frequently with \emph{Bob}, while they interact less frequently, during non-working time. We treat the sequence of emails from one individual to another as an observed dimension, corresponding to a vertex of the dependency graph. The dynamic graph between the four dimensions, aligned with three subintervals, are shown on the top.}
\label{Illustrate}
\end{figure}

There has been growing interests in modeling and understanding temporal dynamics in event occurrences. For instance, modeling customer behaviors and interactions, is crucial for recommendation systems and online social media, to improve the resource allocation and customer experience~\citep{Incentive,MCSN}. 
These   event occurrences usually demonstrate heterogeneous dynamics. On one aspect, individuals usually reciprocate in their interactions with each other (\emph{reciprocity}). For example, if \emph{Alice} sends an email to \emph{Bob}, then \emph{Bob} is more likely to send an email to \emph{Alice} soon afterwards. On the other aspect, long event sequences often exhibit a certain amount of \emph{periodic} trends.  For instance, during working time, individuals are more likely to reciprocate in the email interactions with their colleagues, but such mutually exciting effects will become weaker during non-working time, as illustrated in Fig.~\ref{Illustrate}. 
Temporal point processes (TPPs), such as Hawkes processes (HPs)~\citep{HP}, are in particular well-fitted to capture the \emph{reciprocal}  and \emph{clustering} effects in event dynamics. Nonetheless, the conventional HPs cannot adequately capture the latent state-transition dynamics. 
Recently, neural temporal point processes (neural TPPs) demonstrate a strong  capability in capturing long-range dependencies in event sequences using neural networks~\citep{RMTPP,Xiao2017WassersteinLO,8624540,FullyNN}, attention mechanisms~\citep{SAHP,THP} and neural density estimation~\citep{IntensityFree}. 
These neural TPPs often use all the past events to predict a future event's occurring time, and thus cannot remove the disturbances of the non-contributing events.  
%
To mitigate this defect, some recent works~\citep{NTPPwithGraph,ENTPP} formulate the neural temporal point processes by learning a \emph{static} graph to explicitly capture dependencies among event types. They hence can improve the accuracy in predicting future event time, by removing the non-contributing influences of past events via the learned graph.
%
Nonetheless, the dependencies between event types, may also change over time. Using a \emph{static} graph, the neural TPPs will retrieve a dependency graph averaged over time.  
To fill the gap, some main contributions are made in this paper: (i) We propose to learn a dynamic graph between the event types of an input sequence, 
from a novel variational auto-encoder (VAE) perspective~\citep{VAE,Rezende14}. More specifically, we use regularly-spaced intervals to capture the different states, and 
assume 
\emph{stationary} dynamics within each sub-interval. 
In particular, the dependencies between two event types, is captured using a latent variable, which allows to evolve over sub-intervals. 
We formulate the variational auto-encoder framework, by encoding a latent dynamic graph among event types, from observed sequences. The inter-event waiting times, are decoded using log-normal mixture distributions. 
Via the learned graph, the non-contributing influences of past events, can be effectively removed. 
(ii) The final experiments demonstrate the improved accuracy of the proposed method in predicting event time and types, compared against the existing closely-related methods. The interpretability of the dynamic graph estimated by the proposed method, is demonstrated with both New York Motor Vehicle Collision data. 

\section{Background}
\noindent\textbf{{Multivariate Point Processes}.}
Temporal point processes (TPPs) are concerned with modeling random event sequences in continuous time domain. Let $\mathcal{S}\equiv\{(t_i,v_i)\}_{i=1}^L$ denote a sequence of events, with $t_i\geq 0$ being the timestamp and $v_i\in[1,\ldots,U]$ being the type of $i$-th event. In addition, $\mathcal{H}_t=\{(t_i,v_i)\mid t_i<t,(t_i,v_i)\in\mathcal{S}\}$ denotes the sequence of historical events occurring up to time $t$. 
Multivariate Hawkes processes (MHPs) capture mutually-excitations among event types using the conditional intensity function specified by 
\begin{align}
    \lambda^{*}_v(t) = \mu_v + \sum_{u=1}^U\sum_{\{j:t_j^u<t\}} \alpha_{(v,u)} \exp\left[-\frac{(t-t_j^u)}{\eta_{(v,u)}}\right],\label{MHP}
\end{align}
where $\mu_v$ is the base rate of $v$-th event type, $\alpha_{(v,u)}>0$ captures the instantaneous boost to the intensity due to event $t_j^u$'s arrival, and $\eta_{(v,u)}>0$ determines the influence decay of that event over time.  The stationary condition for MHPs requires $\alpha_{(v,u)}\eta_{(v,u)}<1$. 
In contrast to MHPs, a mutually regressive point process (MRPP)~\citep{MRPP} is to capture both \emph{excitatory} and \emph{inhibitory} effects among event types. These parametric point processes capture a certain form of dependencies on the historical events by designing the conditional intensity functions accordingly. 
Despite being simple and useful, 
these parametric point processes either suffer from certain approximation errors caused by the model misspecifications in practice, or lack the ability to capture long-range dependencies.
To address these limitations, some recent advancements~\citep{RMTPP,FullyNN,IntensityFree,SAHP,THP} combine temporal point processes and deep learning approaches to model complicated dependency structure behind event sequences. At a high level, these neural temporal point processes treat each event as a feature, and encode a sequence of events into an history embedding using various deep learning methods including recurrent neural nets (RNNs), gated recurrent units (GRUs), or long-short term memory (LSTM) nets.  
\citet{RMTPP} uses a recurrent neural net to extract history embedding from observed past events, and then use the history embedding to parameterize its conditional intensity function. The exponential form of this intensity, admits a closed-form integral $\Lambda^{*}(t)=\int_{0}^{t}\lambda^*(s)\mathrm{d}s$, and thus leads to a tractable log-likelihood.  
\citet{LSTM-HP} studied a more sophisticated conditional intensity function, while calculating the log-likelihood involves approximating the integral $\Lambda^{*}(t)$ using Monte Carlo methods. \citet{FullyNN} proposes to model the cumulative conditional intensity function using neural nets, and hence allows to compute the log-likelihood exactly and efficiently. However, sampling using this approach is expensive, and the derived probability density function does not integrate to one. To remedy these issues, \citet{IntensityFree} proposes to {directly} model the inter-event times using normalizing flows. The neural density estimation method~\citep{IntensityFree} not only allows to perform sampling and likelihood computation analytically, but also shows competitive performance in various applications, compared with the other neural TPPs.
\noindent\textbf{{Variational Auto-Encoder}.}
We briefly introduce the definition of variational auto-encoder (VAE), and refer the readers to~\citep{VAE,Rezende14} for more properties. 
VAE is one of the most successful generative models, which allows to straightforwardly sample from the data distribution $p(\mathcal{S})$. It is in particular useful to model high-dimensional data distribution, for which sampling with Markov chain Monte Carlo is notoriously slow. More specifically, we aim at maximizing the data log-likelihood $p(\mathcal{S})$ under the generative process specified by 
$
 p(\mathcal{S})=\int p(\mathcal{S}\mid \mathbf{z},\bm{\theta})p(\mathbf{z})d\mathbf{z},\notag   
$
where $\mathbf{z}$ is the latent variable, and $p(\mathbf{z})$ denotes the prior distribution, and the observation component $p(\mathcal{S}\mid \mathbf{z},\bm{\theta})$ is parameterized by $\bm{\theta}$. 
Under the VAE framework, the posterior distribution $q_{\phi}(\mathbf{z}\mid\mathbf{\mathcal{S}})$ can be defined as $q_{\phi}(\mathbf{z}\mid\mathbf{\mathcal{S}})\equiv\mathcal{N}(\mathbf{z} ; f^{\mu}(\mathcal{S};\phi),f^{\Sigma}(\mathcal{S};\phi))$ where $\mathcal{N}(\cdot)$ refer to a normal distribution, the mean $f^{\mu}(\mathcal{S};\phi)$ and covariance $f^{\Sigma}(\mathcal{S};\phi))$ are parameterized by neural networks with parameter $\phi$.
To learn the model parameters, we maximize the evidence lower bound (ELBO) given by\\ 
$
    \mathcal{L}(\phi,\theta)= \mathsf{E}_{q_{\phi}(\mathbf{z}\mid\mathbf{\mathcal{S}})}\Big[\log p(\mathcal{S} \mid \bm{\theta})\Big]
    -  \mathcal{D}_{\text{KL}}\Big[q_{\phi}(\mathbf{z}\mid\mathbf{\mathcal{S}}) || p_{}(\mathbf{z})\Big],\notag
$
where $\mathcal{D}_{\text{KL}}$ denotes the Kullback–Leibler (KL) divergence. The first term is to make the approximate posterior to produce latent variables $\mathbf{z}$ that can reconstruct data $\mathcal{S}$ as well as possible.
The second term is to match the approximate posterior of the latent variables to the prior distribution of the latent variables.
Using the reparameterization trick~\citep{VAE}, we learn $\phi$ and $\theta$ by maximizing the ELBO using stochastic gradient descent aided by automatic differentiation.
\begin{figure*}[t!]
  \centering
      \includegraphics[width=17.5cm,height=10cm, keepaspectratio]{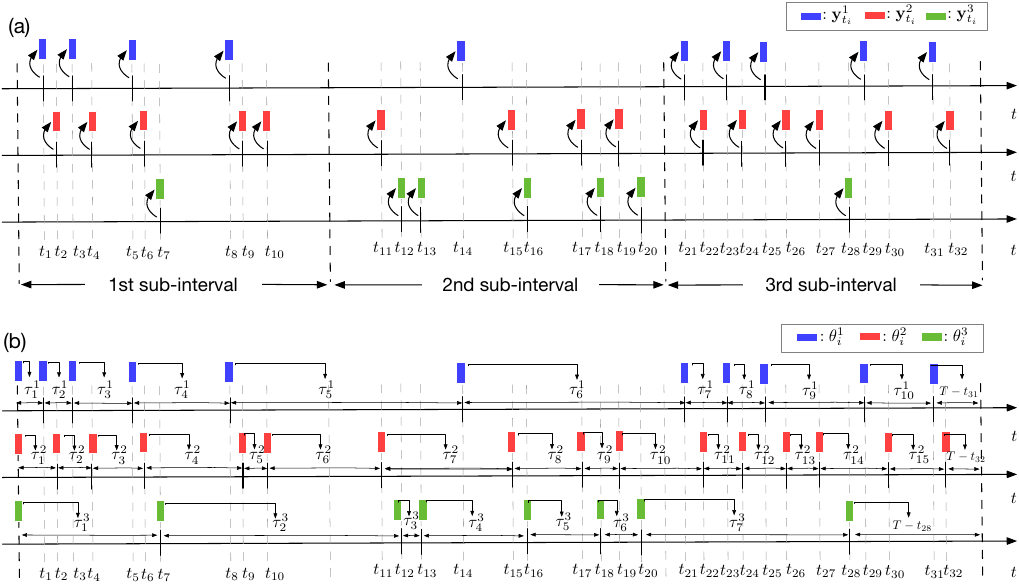}
\caption{(a) The sequence consists of event types and timestamps, which are encoded by the history embedding $\{\mathbf{y}_{t_i}^u\}$ with $u$ being the event type. (b) The decoder captures the inter-event time for a future event, using the log-normal mixture model that is parameterized by the history embeddings. The dynamic graph captures the dependencies between those history embeddings.}
\label{IO}
\end{figure*}

\section{Models}
Given a sequence of events $\mathcal{S}\equiv\{(t_i,v_i)\}_{i=1}^L$, we aim to capture the complicated dependence between event types, using a dynamic graph-structured neural point process. 
The whole time-interval of the sequence is partitioned into $K$ regularly-spaced sub-intervals with $K$ specified a \emph{priori}, to approximately represent the different states. 
We assume 
the latent graph among the event types, is changing over states, but \emph{stationary} within each sub-interval, as illustrated in Fig.~\ref{IO}(a). Specifically, let $[t_k^L,t_k^R)$ stand for the $k$-th sub-interval with $t_k^L$ being the start point and $t_k^R$ being the end point. Let the latent variable $z_{(v,u)}^{k}$ capture the dependence of $v$-th event type on $u$-th event type within $k$-th sub-interval. For ease of exposition, we denote the set of events occurring within the $k$-th sub-interval by\\ $\mathbf{s}^k\equiv\{(t_i,v_i)\mid t_i\in[t_k^L,t_k^R)\}$.  Note that we model the inter-event time for each event type using log-normal mixtures. Hence, we represent the event sequence of $u$-th type by $\mathcal{S}_u=\{(t^u_i,\tau_i^u)\}_{i=1}^{n^u}$, where $t_i^u$ is $i$-th event observed in the sequence of $u$-th event type, $\tau_i^u=t^u_i-t^u_{i-1}$ denotes the corresponding inter-event time, and the total number of events $L=\sum_{u=1}^{U}n^u$.  
Next we shall explain each component of the variational autoencoder in the following subsections.

\noindent\textbf{{Prior}.}
We assume that the dependency graph among event types, evolves over sub-intervals. Hence, we use an autoregressive model to capture the the prior probabilities of the latent variables $\{\mathbf{z}_{(v,u)}^k\}_{v,u,k}$. More specifically, the prior distribution of $\mathbf{z}_{(v,u)}^k$ depends on its previous state $\mathbf{z}_{(v,u)}^{k-1}$ and the event sequence up to time $t_k^L$ (all the events in first $k-1$ sub-intervals):
\begin{align}
    p_{\phi}(\mathbf{z}\mid \mathcal{S})\equiv \prod_{k=1}^{K}p_{\phi}(\mathbf{z}^k\mid \mathbf{z}^{1:k-1},\mathbf{s}^{1:k}).\notag
\end{align}

\begin{figure}[t]
  \centering
      \includegraphics[width=8.3cm,height=10cm, keepaspectratio]{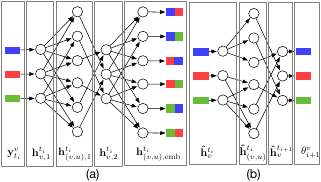}
\caption{(a) A fully-connected GNN is used to transform event embeddings into relation embeddings between event types at each timestamp. (b) A GRNN is used to transform the past influences of those related event types into the current embedding according to the dynamic graphs.}
\label{ENC}
\end{figure}

The prior component is specified as follows: for each event of $v$-th type $(t_i^v,m_i^v)$, where $m_i^v$ represents the auxiliary event mark if available, we embed $t_i^v$ and $m_i^v$ into a fixed-dimensional vector $\mathbf{y}_{v}^{t_i}\in\mathbb{R}^D$. Then, we pass the event embedding $\mathbf{y}_{v}^{t_i}$ through a fully-connected graph neural network (GNN) to obtain the relation embedding $\mathbf{h}^{t_i}_{(v,u),\text{emb}}$ between event types $v$ and $u$:
\begin{alignat*}{3}
& &\mathbf{h}_{v,1}^{t_i}&=f^1_{\text{emb}}(\mathbf{y}_v^{t_i}),\\
&v\rightarrow e: &\mathbf{h}_{(v,u),1}^{t_i}&=f_{\text{e}}^1([\mathbf{h}_{v,1}^{t_i},\mathbf{h}_{u,1}^{t_i}]),\\
&e\rightarrow v: &\mathbf{h}_{v,2}^{t_i}&=f_v^1\Big(\sum_{u\neq v}\mathbf{h}_{(v,u),1}^{t_i}\Big),\\
&v\rightarrow e:&\mathbf{h}_{(v,u),\text{emb}}^{t_i}&=f_{\text{e}}^2([\mathbf{h}_{v,2}^{t_i},\mathbf{h}_{u,2}^{t_i}]),
\end{alignat*}
where $f(\cdot)$ denotes a multilayer perceptron (MLP) for each layer of the GNN, $\mathbf{h}_{v,\ell}^{t_i}$ and $\mathbf{h}_{(v,u),\ell}^{t_i}$ represent the node-wise and edge-wise hidden states of the $\ell$-th intermediate layer, respectively. 
The final output of the GNN $\mathbf{h}_{(v,u),\text{emb}}^{t_i}$ models the relation at time $t_i$. The GNN architecture is illustrated in Fig.~\ref{ENC}(a).

We need to concatenate all the relation variable, $\{ \mathbf{h}_{(v,u),\text{emb}}^{t_i}\}\ \text{for}\ t_i\in[t^L_k,t^R_k)$, and transform them into the relation state $\mathbf{h}_{(v,u),\text{emb}}^{k}$ of $k$-th sub-interval, using a MLP:  
\begin{alignat*}{3}
& &\mathbf{h}_{(v,u),\text{emb}}^{k}&=f^2_{\text{emb}}([\mathbf{h}_{(v,u),\text{emb}}^{t_i}])\quad\ \text{for}\ t_i\in[t^L_k,t^R_k).
\end{alignat*}

A forward recurrent neural network (RNN) is used to capture the dependence of an relation state $\mathbf{h}_{(v,u),\text{fwd}}^{k}$ on its current embedding $\mathbf{h}_{(v,u),\text{emb}}^{k}$, and its previous state $\mathbf{h}_{(v,u),\text{fwd}}^{k-1}$: 
\begin{align}
    \mathbf{h}_{(v,u),\text{fwd}}^{k}&=\text{RNN}_{\text{fwd}}(\mathbf{h}_{(v,u),\text{emb}}^{k},\mathbf{h}_{(v,u),\text{fwd}}^{k-1}).\notag
\end{align}

Finally, we encode $\mathbf{h}_{(v,u),\text{fwd}}^{k}$ into the logits of the prior distribution for $\mathbf{z}_{(v,u)}^k$, using a MLP:
\begin{align}
    p_{\phi}(\mathbf{z}_{(v,u)}^k\mid \mathbf{z}^{1:k-1},\mathbf{s}^{1:k})&=\text{softmax}(f_{\text{prior}}(\mathbf{h}_{(v,u),\text{fwd}}^{k})).\notag
\end{align}

\begin{figure}[t]
  \centering
      \includegraphics[width=8.3cm,height=10cm, keepaspectratio]{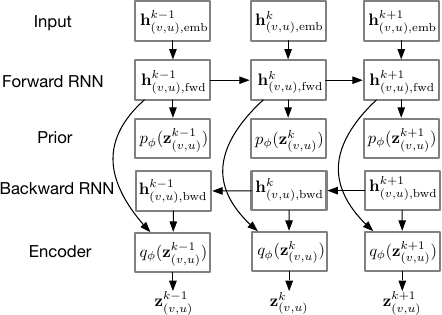}
\caption{The relation embeddings are fed into a forward RNN and a backward RNN to encode the influences from the past and future relations, respectively.}
\label{FB_GNN}
\end{figure}

Fig.~\ref{FB_GNN} shows the prior distribution built upon the foward RNN.

\noindent\textbf{{Encoder}.}
The posterior distribution of the latent variables $q_{\phi}(\mathbf{z\mid\mathcal{S}})$ depends on both the past and future events: 
\begin{align}
    q_{\phi}(\mathbf{z}\mid \mathcal{S})\equiv \prod_{k=1}^{K}q_{\phi}(\mathbf{z}^k\mid \mathcal{S}).\notag
\end{align}

Hence, the encoder is designed to approximate the distribution of the relation variables using the whole event sequence. To that end, a backward GNN is used to propagate the hidden states $\mathbf{h}_{(v,u),\text{bwd}}^{k}$ reversely:
\begin{align}
    \mathbf{h}_{(v,u),\text{bwd}}^{k}&=\text{RNN}_{\text{bwd}}(\mathbf{h}_{(v,u),\text{emb}}^{k},\mathbf{h}_{(v,u),\text{bwd}}^{k+1}).\notag
\end{align}

Finally, we concatenate both the forward state $\mathbf{h}_{(v,u),\text{fwd}}^{k}$ and the backward state $\mathbf{h}_{(v,u),\text{bwd}}^{k}$, and transform them into the logits of the approximate posterior, using a MLP: 
\begin{align}
    q_{\phi}(\mathbf{z}_{(v,u)}^k\mid \mathcal{S})&=\text{softmax}(f_{\text{enc}}([\mathbf{h}_{(v,u),\text{bwd}}^{k},\mathbf{h}_{(v,u),\text{fwd}}^{k}])).\notag
\end{align}

Note that the prior and encoder share parameters, and thus the parameters of the two components are denoted by $\phi$.

\noindent\textbf{{Decoder}.}
The role of the decoder is to predict the inter-event times $\{\tau_i^u\}_{i=1}^{n_u}$ for each event type $u$. 
In particular, we capture the latent dynamics $\{\mathbf{\hat{h}}_{v}^{t_i}\}$ behind these inter-event times using a graph recurrent neural network (GRNN) specified by 
\begin{alignat*}{3}
&v\rightarrow e: &\mathbf{\hat{h}}_{(v,u)}^{t_i}&=z_{(v,u)}^k f_{\text{e}}^1([\mathbf{\hat{h}}_{v}^{t_i},\mathbf{\hat{h}}_{u}^{t_i}]),\quad\ \text{for}\ t_i\in[t^L_k,t^R_k),\\
&e\rightarrow v: &{\mathbf{\tilde{h}}}_{v}^{t_i}&= \sum_{u\neq v}\mathbf{\hat{h}}_{(v,u)}^{t_i},  \\
& &\mathbf{\hat{h}}_{v}^{t_{i+1}}&=\text{GRU}({\mathbf{\tilde{h}}}_{v}^{t_i},{\mathbf{\hat{h}}}_{v}^{t_i}), 
\end{alignat*}
where $z_{(v,u)}^k$ determines how the $u$-th event type $\mathbf{\hat{h}}_{u}^{t_i}$ influences the $v$-th event type $\mathbf{\hat{h}}_{v}^{t_{i+1}}$ through the relation $z_{(v,u)}^k$ at time $t_{i+1}$. The latent embedding $\mathbf{\hat{h}}_{v}^{t_i}$ itself evolves over time, using a gated recurrent unit (GRU).


Given the dynamic embeddings $\{\mathbf{\hat{h}}_{v}^{t_i}\}$, we model the inter-event time $p(\tau_i^u)$ using a log-normal mixture model~\cite{IntensityFree}, 
\begin{align}
    p(\tau\mid \bm{\omega},\bm{\mu},\bm{\sigma})=\sum_{c=1}^{C} {\omega}_c \frac{1}{\tau{\sigma}_c\sqrt{2\pi}}\exp\Big(-\frac{(\log\tau-{\mu}_c)^2}{2{\sigma}_c^2}\Big),\notag
\end{align}
where $\omega_c,\mu_c,\sigma_c$ represent the mixture weights, the mean and the standard deviations of $c$-th mixture component, respectively.
In particular, the parameters of the distribution for each inter-event time $\tau_i^u$ is constructed as
\begin{align}
    \bm{\omega}_i^{u}&=\text{softmax}(V_{\bm{\omega}}\mathbf{\hat{h}}_{u}^{t_i}+\bm{\beta_{\omega}}),\quad\
    \bm{\sigma}_i^{u}=\text{exp}(V_{\bm{\sigma}}\mathbf{\hat{h}}_{u}^{t_i}+\bm{\beta_{\sigma}}),\notag\\
    \bm{\mu}_i^{u}&=V_{\bm{\mu}}\mathbf{\hat{h}}_{u}^{t_i}+\bm{\beta_{\mu}},\notag
\end{align}
where $\{V_{\bm{\omega}},V_{\bm{\sigma}},V_{\bm{\mu}},\bm{\beta_{\omega}},\bm{\beta_{\sigma}},\bm{\beta_{\mu}}\}$ refer to the learnable parameters. We use $\texttt{softmax}$ and $\texttt{exp}$ transformations to impose the sum-to-one and positive constraints on the distribution parameters accordingly. Fig.~\ref{ENC}(b) presents the GNN architecture used to construct the parameters of the log-normal mixtures in the decoder part. 
We assume the inter-event time $\tau_i^u$ is conditionally independent of the past events, given the model parameters. Hence, the distribution of the inter-event time under the decoder, factorizes as
\begin{align}
p_{\theta}(\bm{\tau} \mid \mathbf{z}) = \prod_{u=1}^{U}\prod_{i=1}^{n^u} p(\tau_{i}^u\mid \bm{\theta}_{i}^{u}).\notag
\end{align}

The decoder part of the VAE framework is illustrated in Fig.~\ref{IO} (b). Hence, we can naturally make the next event time prediction using
\begin{align}
    \hat{t}^u_{i+1} &= t^u_{i} + \int_{0}^{\infty}\tau p(\tau_{i+1}^u\mid \bm{\theta}_{i+1}^{u})\mathrm{d}\tau.\notag
\end{align}


\noindent\textbf{{Training}.}
We next explain how to learn the parameters of the VAE framework for dynamic graph-structured neural point processes. 
The event sequences $\mathcal{S}$ are passed through the GNN in the encoder, to obtain the relation embedding $\mathbf{h}^{t_i}_{(v,u),\text{emb}}$ for all the timestamps $\{t_i\}_{i=1}^L$ and each pair of two event types $(v,u)$. 
We then concatenate all the relation embeddings, and transform them into a relation state $\mathbf{h}^{k}_{(v,u),\text{emb}}$ for each sub-interval $k$.
The relational states $\{\mathbf{h}^{k}_{(v,u),\text{emb}}\}$ are fed into the forward and backward RNNs to compute the prior distribution $p_{\phi}(\mathbf{z}\mid \mathcal{S})$ and posterior distribution $q_{\phi}(\mathbf{z}\mid \mathcal{S})$. We then sample $\{\mathbf{z}_{(v,u)}^k\}$ from the concrete reparameterizable approximation of the posterior distribution. 
The hidden states $\{\mathbf{\hat{h}}_{v}^{t_i}\}$ evolve through a GRNN, in which the messages can only pass through the non-zero edges hinted by $\{\mathbf{z}_{(v,u)}^k\}$. These hidden states $\{\mathbf{\hat{h}}_{v}^{t_i}\}$ are used to parameterize the log-normal mixture distribution for the inter-event times. To learn the model parameters, we calculate the evidence lower bound (ELBO) as
\begin{align}
    \mathcal{L}^{\mathrm{ELBO}}(\phi,\theta)=\ \mathsf{E}_{q_{\phi}(\mathbf{z}^{k}\mid\mathbf{\mathcal{S}})}\Big[\sum_{u=1}^{U}\sum_{i=1}^{n^u} \log p(\tau_{i}^u\mid \bm{\theta}_{i}^{u})\Big] \label{ELBO}\\
    - \sum_{k=1}^K \mathcal{D}_{\text{KL}}\Big[q_{\phi}(\mathbf{z}^{k}\mid\mathbf{\mathcal{S}}) || p_{\phi}(\mathbf{z}^k\mid \mathbf{z}^{1:k-1},\mathbf{s}^{1:k})\Big].\notag
\end{align}

As we draw samples $\{\mathbf{z}_{(v,u)}^k\}$ using an reparameterizable approximation, we can calculate gradients using backpropagation and optimize the ELBO. Hereafter, we denote the proposed model as variational autoencoder temporal point process (VAETPP).

\section{Related Work}
\citet{SN1980,SN03} proposed to capture the non-stationary network dynamics using continuous-time Markov chains. 
\noindent\textbf{Graph-Structured Temporal Point Processes.}
\citet{PGEM} proposed a proximal graphical event model to infer the relationships among event types, and thus assumes that the occurrence of an event only depends on the occurring of its parents shortly before. 
\citet{GHP} developed a geometric Hawkes process to capture correlations among multiple point processes using graph convolutional neural networks although the inferred graph is {undirected}. 
\citet{THP} developed a graph-structured transformer Hawkes process between multiple point processes. It assumes that each point process is associated with a vertex of a \emph{static} graph, and model the dependency among those point processes by incorporating that graph into the attention module design. \citet{NTPPwithGraph} formulated a graph-structured neural temporal point process (NTPP) by sampling a latent graph using a generator. The model parameters of the graph-structured NTPP and its associated graph, can be simultaneously optimized using an efficient bi-level programming. \citet{ENTPP} recently also developed a variational framework for graph-structured neural point processes by generating a latent graph using intra-type history embeddings. The latent graph is then used to govern the message passing between intra-type embedding to decode the type-wise conditional intensity function. \citet{ijcai2021p469} proposed to learn a static conditional graph  between event types, by incorporating prior knowledge. 
Additionally, \citet{LGPP} proposed to learn the latent graph structure among events via a probabilistic model. \citet{GBTPP} tries to model event propagation using a graph-biased temporal point process, with a graph specified by the following relationships among individuals on social media.  \citet{VAEPP} studied a variational auto-encoder framework for modeling event sequences but haven't consider capturing latent graphs among dimensions. 
\noindent\textbf{Dynamic Latent Graphs behind Time Series.} 
\citet{NRI} developed an variational autoencoder to learn a latent \emph{static} graph among the entities that interact in a physical dynamical system. Following this success, \citet{dNRI} used a sequential latent variable model to infer dynamic latent graphs among the entities from the discrete-time observations. Although \citet{dNRI} and the proposed method both aim to learn dynamic graphs behind multivariate sequential observations, the main differences lie in that our input is \emph{asynchronous} event sequences, for which we need to determine the regularly-spaced sub-intervals, and simultaneously learn a dynamic graph for each sub-interval. An interesting yet challenging direction, is to automatically infer the regularly-spaced time intervals and the dynamic graph structure, which we leave to future research. 

\section{Experiments}
The proposed variational autoencoder temporal point process, is evaluated on the task of event time and type prediction. We used four 
real-world data to demonstrate the proposed method, compared with existing related methods. 
\noindent\textbf{New York Motor Vehicle Collisions(NYMVC):} 
This data contains a collection of vehicle collision events occurring at New York city since April, 2014. Each crash event $(t_i,v_i)$ records a motor vehicle collision occurring in district $v_i$, at time $t_i$. Specifically, during peak periods, a vehicle collision may give rise to a sequence of collisions in the same or nearby districts, in short time. Hence, it is well-fitted to model and predict the occurrences of these events using multivariate point processes. In addition, the influence relations among the districts may change over time as the aforementioned triggering effects become weaker during night time, compared to day time. We created each event sequence using the motor vehicle collision records between 8:00 and 23:00, and treated each three hours as a sub-interval. We considered the five districts, \emph{Manhattan}, \emph{Brooklyn}, \emph{Bronx}, \emph{Queens} and \emph{Staten Island} as the event types.
\begin{table*}[h]
\centering
\small
\begin{tabular}{lcccc}\hline
Methods & {MathOF} & {AskUbuntu} & {SuperUser} & {NYMVC} \\ \hline 
Exponential &  $2.549\pm0.074$ & $2.584\pm0.029$ & $2.517\pm0.018$ & $2.474\pm0.043$  \\ 
RMTPP &  $1.912\pm0.087$ & $1.981\pm0.014$ & $2.025\pm0.054$ & $1.944\pm0.012$  \\ 
FullyNN &  $1.652\pm0.062$ & $1.884\pm0.073$ & $1.777\pm0.023$ & $1.473\pm0.024$   \\ 
LogNormMix &  ${-0.859\pm0.121}$ & $0.303\pm0.037$ & $-0.868\pm0.018$ & ${-2.578\pm0.032}$   \\ 
THP &  ${-2.531\pm0.024}$ & $-2.235\pm0.028$ & $-2.349\pm0.051$ & ${-1.889\pm0.037}  $   \\ 
VAETPP (static) & ${-2.632\pm0.028}$ & ${-2.312\pm0.026}$ & ${-2.466\pm0.021}$ & ${-2.016\pm0.032}$   \\ 
VAETPP & $\mathbf{-3.501\pm0.068}$ & $\mathbf{-2.867\pm0.032}$ & $\mathbf{-3.812\pm0.057}$ & $\mathbf{-5.952\pm0.046}$   \\ \hline
\end{tabular}
\caption{\label{NLL} {Negative log-likelihood for inter-event time prediction on the real-world data.}}
\end{table*}
\begin{table}[h]
\centering
\small
\begin{tabular}{lccc}\hline
Datasets & {$\#$ sequences} & {$\#$ events} & {$\#$ types} \\ \hline 
MathOF &  $1453$ & $590836$ & $15$   \\
AskUbuntu &  $1561$ & $65960$ & $11$   \\
SuperUser &  $1240$ & $84627$ & $10$   \\
NYMVC &  $2000$ & $863624$ & $5$   \\ \hline
\end{tabular}
\caption{\label{SSData} {Statistics of the datasets.}}
\end{table} 
\noindent\textbf{Stack Exchange Data:} 
The three stack exchange data from different sources, are included in the experiments: \emph{MathOF}, \emph{AskUbuntu}, and \emph{SuperUser}. The stack exchange data consists of various interactions among the participant. Each event $(v_i,u_i,t_i)$ means that at timestamp $t_i$, user $v_i$ may post an answer or comment to $u_i$'s questions or comments. These interaction events among users usually exhibit a certain amount of \emph{clustering} effects and periodic trends. For instance, some questions about popular technology, may quickly lead to a lot of answers or comments from the others who share similar interests. Additionally, these triggering effects demonstrate periodic trends: users are more inclined to response to technical topics during working days, than weekends/holidays. We consider the user who makes the action toward the other as the event type. 
 Hence, we derived each sequence from events occurring within one week, and consider each day as a sub-interval. These datasets are detailed in Table~\ref{SSData}.

\begin{table*}[h]
\centering
\small
\begin{tabular}{lcccc}\hline
Methods & {MathOF} & {AskUbuntu} & {SuperUser} & {NYMVC} \\ \hline 
RMTPP &  $0.952\pm0.008$ & $0.983\pm0.025$  & $1.103\pm0.068$ & $1.135\pm0.095$  \\ 
LogNormMix &  ${0.673\pm0.082}$ & $0.969\pm0.043$ & $0.708\pm0.098$ & ${0.798\pm0.038}$   \\ 
THP &  ${0.693\pm0.024}$ & $0.791\pm0.057$ & $0.779\pm0.032$ &  ${0.859\pm0.027}$   \\ 
VAETPP(static) & ${0.632\pm0.020}$ & ${0.812\pm0.032}$ & ${0.788\pm0.038}$ & ${0.864\pm0.046}$   \\ 
VAETPP & $\mathbf{0.569\pm0.018}$ & $\mathbf{0.642\pm0.012}$ & $\mathbf{0.674\pm0.046}$ & $\mathbf{0.775\pm0.064}$   \\ \hline
\end{tabular}
\caption{\label{RMSE} {Root mean square error (RMSE) comparison for event time prediction.}}
\end{table*}
\begin{table*}[h]
\centering
\small
\begin{tabular}{lcccc}\hline
Methods & {MathOF} & {AskUbuntu} & {SuperUser} & {NYMVC} \\ \hline 
RMTPP &  $0.154\pm0.022$ & $0.189\pm0.021$ & $0.208\pm0.008$ & $0.251\pm0.025$  \\ 
LogNormMix &  ${0.206\pm0.027}$ & $0.225\pm0.011$ & $0.235\pm0.013$ & ${0.276\pm0.038}$   \\ 
THP &  ${0.242\pm0.019}$ & $0.261\pm0.016$ & $0.248\pm0.006$ &  ${0.294\pm0.028}$   \\ 
VAETPP(static) & ${0.286\pm0.031}$ & ${0.286\pm0.022}$ & ${0.256\pm0.018}$ & ${0.278\pm0.026}$   \\ 
VAETPP & $\mathbf{0.321\pm0.016}$ & $\mathbf{0.318\pm0.012}$ & $\mathbf{0.288\pm0.018}$ & $\mathbf{0.301\pm0.022}$   \\ \hline
\end{tabular}
\caption{\label{ACC} {Event type prediction accuracy comparison.}}
\end{table*}

\noindent\textbf{Experimental Setup.} 
We compared the ability of the models in predicting the inter-event times $\tau_i^u$ using the historical events $\mathcal{H}_{t_i^u}$, for each event type $u\in [1,\ldots,U]$, as illustrated in Fig.~\ref{IO}(b). Each real-world data is split into multiple event sequences. 
For each real-world data, we choose the $60\%$ of the sequences for training, $20\%$ for validation, and $20\%$ for testing. For training, we maximize the ELBO in Eq.~\ref{ELBO} for the proposed model, and the expected log-likelihood for the other models. 
With the learned parameters, we measure the predictive performance of each model, using its obtained negative log-likelihood (NLL) on the validation set. 
Hence, the model configuration that achieves best predictive performance, can be chose using the validation set. Finally, the NLL loss on the test set, are used to compare the ability of the models in predicting inter-event times. We report the results averaged over ten random training/validation/testing splits. 
For the developed VAETPP, the dimension of the input embedding $\mathbf{y}_{t_i}^u$ is $64$. For the fully-connected GNN of the encoder, $f^1_{emb}, f_e^1,f_v^1$, and $f_e^1$ are two-layer MLPs with 64 units for each layer and Exponential Linear Unit (ELU) activations. We use $f_{emb}^2$ to transform the concatenated hidden states within each sub-interval into one hidden state, and thus parameterize $f_{emb}^2$ using a one-layer MLP with 64 hidden units and ReLU activations.  
Both the forward RNN and backward RNN have 64 hidden units. We parameterize $f_{\text{prior}}$ and $f_{\text{enc}}$ by a one-layer MLP with 64 hidden units and Rectified Linear Unit (ReLU) activations. We set the number of edge types of the dynamic graph among event types to be two, and specify the first edge type to indicate no dependency. For the decoder part, we parameterize $f_e^1$ using a separate two-layer MLP with 64 hidden/output units, for each of the two edge types. The GRU has 64 hidden units. 
We chose the number of mixing components used in the log-normal mixture distribution for the VAETPP, using the validation data. In the experiments, we used $16$ mixing components in the experiments. We also consider restricting the VAETPP with a static latent graph, and denote it as VAETPP (static), to validate the importance of learning dynamic graphs in capturing period trends in event sequences. 
We compared the proposed method against the following baselines: 
\textbf{Exponential.} The conditional intensity function of the constant intensity model~\citep{Exponential} is defined as $\lambda^{*}(t_i)=\exp{(\mathbf{v}^{\mathrm T}\mathbf{h}_i+\mathbf{b})}$, in which $\mathbf{h}_i$ denotes the event history embedding learned by a RNN,  $\mathbf{v}$ and $\mathbf{b}$ refer to the model parameters. The probability density function (PDF) of the constant intensity model is an exponential distribution, as given by $p^{*}(\tau)=\gamma\exp{(-\gamma)}$, where $\gamma=\exp{(\mathbf{v}^{\mathrm T}\mathbf{h}_i+\mathbf{b})}$.
\textbf{Recurrent Marked Temporal Point Processes(RMTPP)} \citep{RMTPP}. The method encodes past events into historical embeddings using a RNN, and models exponential-distributed conditional intensity.
\textbf{Fully Neural Networks (FullyNN)} \citep{FullyNN}. It captures the cumulative distributions of inter-event times using a neural network. 
\textbf{Log Normal Mixture (LogNormMix)} \citep{IntensityFree}. The method encodes the event history into embedding vectors with a RNN, and decodes the inter-waiting time with a log-normal mixture distribution. 
\textbf{Transformer Hawkes Process (THP)} \citep{THP}. It leverages the self-attention mechanism to capture long-term dependencies in observed event sequences. 

\noindent\textbf{Negative Log-likelihood Comparison.} Table~\ref{NLL} compares the negative log-likelihood loss of all the methods in modeling inter-event times. As expected, the LogNormMix admits more flexibility compared with simple model using unimodal distributions (Gompertz/RMTPP, Exponential), and thus shows improved performance by large margins. Transformer Hawkes process (THP) can effectively learn long-range dependencies among events, and thus achieves lower NLL loss. The proposed VAETPP not only can capture the complicated inter-event time distribution using a log-normal mixture decoder. It can also improve the inter-event time prediction in further by effectively removing the non-contributing effects of irrelevant past events, via the dynamic dependency graph. Hence, the proposed VAETPP achieves the best NLL loss values consistently on all the datasets.
\begin{figure}[ht]
\centering
 \includegraphics[width=8.2cm,height=9cm, keepaspectratio]{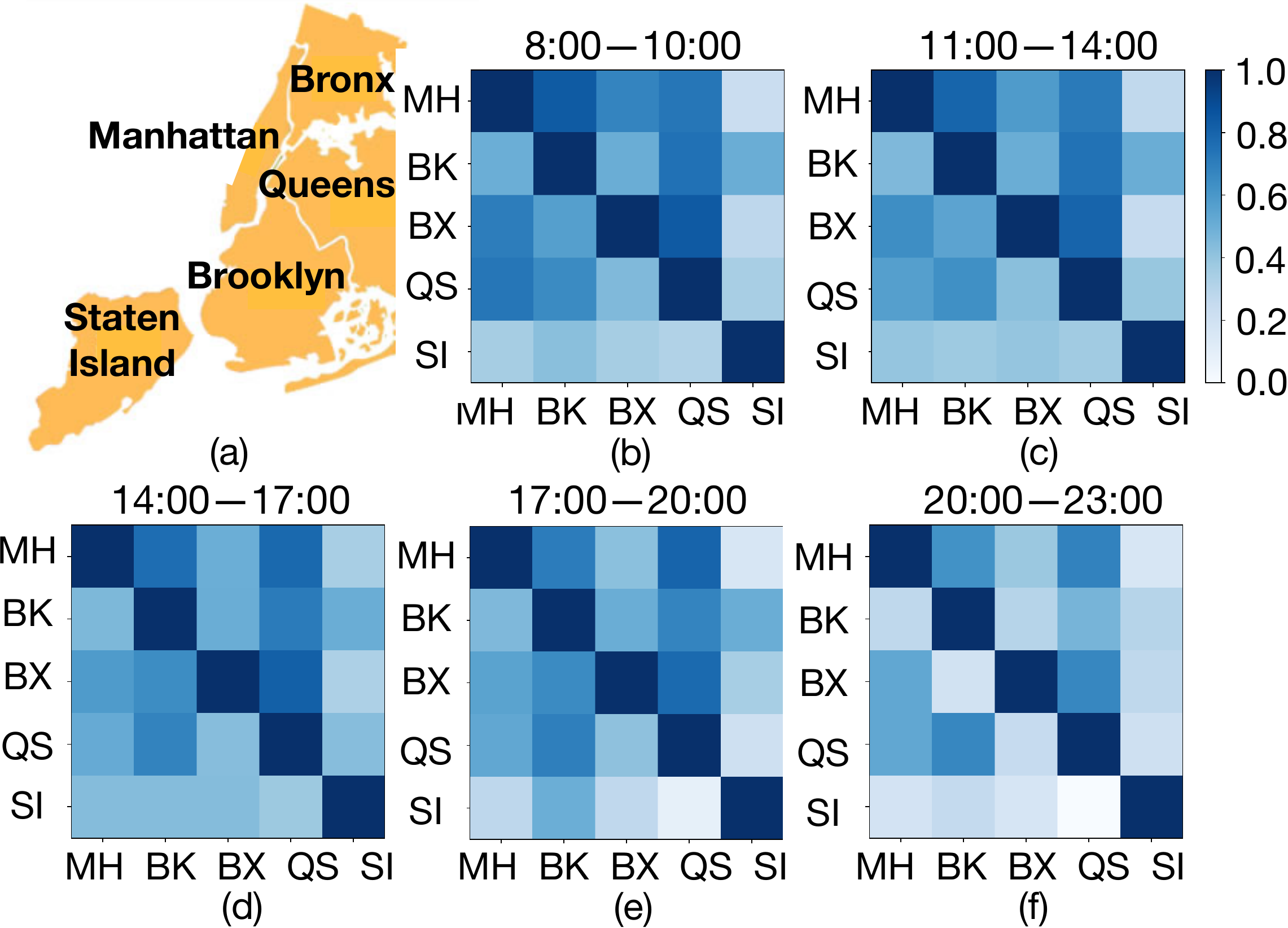}
    \caption{(a) shows the relative locations of the five boroughs of New York city, Manhattan (MH), Bronx (BX), Brooklyn (BK), Queens (QS), Staten Island (SI); (b-f) show the latent dynamic graph between the five boroughs over five time-intervals, estimated by the VAETPP.}\label{NYC}
\end{figure}

\noindent\textbf{Event Prediction Comparison.}
We also consider the tasks of event time and type prediction in the experiments. In particular, following~\citep{THP}, 
we make the next event time prediction using a linear predictor as 
$
    \hat{t}_{i+1}^u = \mathbf{W}^{\text{time}}\bm{\theta}_{i}^u,\notag
$
where $\bm{\theta}_{i}^u$ is the history embedding updated by the VAETPP after observing of $i$-th event of $u$-th type, and  $\mathbf{W}^{\text{time}}\in\mathbb{R}^{1\times D}$ denotes the event time predictor's parameter.  
The next event type prediction is  
\begin{align}
    \hat{\mathbf{p}}_{i+1} =\texttt{softmax}(\mathbf{W}^{\text{type}}\bm{\theta}_{i}^u),  \ \ 
    \hat{m}_{i+1}^u = \arg\max_{j} \hat{\mathbf{p}}_{i+1}(j),\notag
\end{align}
 where $\mathbf{W}^{\text{type}}\in\mathbb{R}^{J\times D}$ denotes the event type predictor's parameter, and $\hat{\mathbf{p}}_{i+1}(j)$ refers to the $j$-th entry of $\hat{\mathbf{p}}_{i+1}\in\mathbb{R}^{\scriptscriptstyle J}$.
The loss functions for event time and type prediction are defined as
\begin{align}
\mathcal{\widetilde L}(\mathcal{S};\theta) =\sum_u\sum_{i=1}^{n_u} (t_i^u - \hat{t}_i^u)^2,
\mathcal{\widehat L}(\mathcal{S};\theta) =-\sum_{j=2}^L \mathbf{m}_j^{\mathrm T}\log(\mathbf{\hat p}_j),\notag
\end{align}
 respectively, where $\mathbf{m}_j$ is the one-hot encoding for
the type of $j$-th event. 
To learn the parameters of the event time and type predictor, we consider minimizing a composite loss function as\\
$
\min_{\phi,\theta} -\mathcal{L}^{\mathrm{ELBO}}(\mathcal{S};\phi,\theta) +
\mathcal{\widetilde L}(\mathcal{S};\theta) + \mathcal{\widehat L}(\mathcal{S};\theta),\notag
$
where $\mathcal{L}^{\mathrm{ELBO}}(\mathcal{S};\phi,\theta)$ is the evidence lower bound derived in Eq.\ref{ELBO}.  
We used the training data to learn the model parameters, and choose the best configuration according to the predictive performance on the validation set. Finally, we evaluated model performance on the test set. Specifically, we predicted each held-out event $(t_j,m_j)$ given its history. We evaluated event type prediction by accuracy and event time prediction by Root Mean Square Error (RMSE). Tab.~\ref{RMSE} and~\ref{ACC} shows the results for event time and type prediction, respectively. Our VAETPP outperforms the baselines in predicting event time and types on all the data.

\noindent\textbf{Model Interpretability.}
Fig.~\ref{NYC}(a) shows the relative location of the five boroughs of New York city: Manhattan, Brooklyn, Bronx, Queens, Staten Island. The latent dynamic graphs for the five time-intervals, are plotted in Fig.~\ref{NYC}(b-f). From the results, we find that the influences between Manhattan, Brooklyn, Bronx and Queens, are much stronger, compared with influences between these four areas and Staten Island. These influences between all the areas, gradually become weaker during night time, compared against day time. The results not only show the high model interpretability, but also explains why VAETPP obtains better prediction accuracy by effectively removing non-contributing historical events' influences via its estimated graphs.

\section{Conclusion}
We have presented a novel variational auto-encoder for modeling asynchronous event sequences. To capture the \emph{periodic} trends behind long sequences, we use regularly-spaced intervals to capture the different states behind the sequences, and assume stationary dynamics within each sub-interval. The dependency structure between event types, is captured using latent variables, which allow to be evolving over time, to capture the time-varying graphs. Hence, the proposed model can effectively remove the influences from the irrelevant past event types, and achieves better accuracy in predicting inter-event times and types, compared with other neural point processes. We plan to generalize the work to capture non-stationary network dynamics~\cite{AAAI-18,ICML-18,UAI-20,SDM-23} in the future research.

\section{Acknowledgments}
The work is partially funded by Shenzhen Science and Technology Program (JCYJ20210324120011032) and Shenzhen Institute of Artificial Intelligence and Robotics for Society.
%
%

\bibliography{main}

\end{document}